# A New Type of Foundation Model Based on Recordings of People's Emotions and Physiology

David Gamez,[1] Dionis Barcari[1] and Aliya Grig[2]

1. Department of Computer Science, Middlesex University, London, UK
2. Evolwe, 251 Little Falls Drive, Wilmington, Delaware 19808, United States

**Abstract.** Foundation models have had a big impact in recent years and billions of dollars are being invested in them in the current AI boom. The more popular ones, such as Chat-GPT, are trained on large amounts of data from the Internet, and then reinforcement learning, RAG, prompt engineering and cognitive modelling are used to fine-tune and augment their behavior. This technology has been used to create models of individual people, such as Caryn Marjorie. However, these chatbots are not based on people's actual emotional and physiological responses to their environment, so they are, at best, surface-level approximations to the characters they are imitating. This paper describes how a new type of foundation model – a first-person foundation model - could be created from recordings of what a person sees and hears as well as their emotional and physiological reactions to these stimuli. A first-person foundation model would map environmental stimuli to a person's emotional and physiological states, and map a person's emotional and physiological states to their behavior. First-person foundation models have many exciting applications, including a new type of recommendation engine, personal assistants, generative adversarial networks, dating and recruitment. To obtain training data for a first-person foundation model, we have developed a recording rig that captures what the wearer is seeing and hearing as well as their emotional and physiological states. This novel source of data could help to address the shortage of new data for building the next generation of foundation models.

**Keywords:** Foundation models, Large language models, Emotion and decision making, First-person recorder, Personality modelling, Artificial intelligence, recommendation, cold-start problem, digital twin.

## 1 Introduction

Foundation models have had a big impact in recent years and billions of dollars are being invested in them in the current AI boom. The more popular ones, such as Llama, Chat-GPT and Dall-E, can generate plausible text and images in response to text and image prompts. These models are trained on large amounts of text and image data scraped from the Internet - often accessed through the Common Crawl repository. Researchers and technology companies are starting to realize that this data source could become exhausted soon, so they are looking for new sources of data to train the next generation of foundation models [1] [2].

After initial training the behavior of foundation models can be fine-tuned through a combination of reinforcement learning, retrieval augmented generation (RAG) and prompt engineering. These techniques have been used to create fairly plausible imitations of individual people, such as Marilyn Monroe (https://www.soulmachines.com/) and Caryn Marjorie (https://caryn.ai). However, these chatbots are not based on people's actual physiological and emotional responses, so they are, at best, a surface level approximation to the characters they are imitating.

To address these issues, this paper outlines how a new type of foundation model - a first person-foundation model (FPFM) - could be created from recordings of what a person sees and hears as well as their emotional and physiological reactions to these stimuli. A FPFM would map environmental stimuli to a person's emotional and physiological reactions and map a person's emotional and physiological states to their behavior. First-person foundation models have many exciting applications, including a new type of recommendation engine, personal assistants, generative adversarial networks, dating and recruitment. To acquire data for training a FPFM we have developed a recording rig that captures what the wearer is seeing and hearing as well as their skin conductance (GSR), facial expression and brain state (14 channel EEG). AI algorithms are used to process this data into a rich picture of the environment and internal states of the subject.

The first part of this paper gives background information about foundation models, personality modelling with foundation models, the role of emotional and physiological states in decision-making, and recommendation systems. Section 3 suggests how we could develop first-person foundation models and outlines some of their use cases. It also explains how fine-tuning, RAG and prompt engineering could be combined with FPFMs to create digital twins of people that could be shared with third parties to support recommendation, advertising, and other applications. Section 4 describes the recording rig that we have developed to capture first-person training data for FPFMs, and the paper concludes with some reflections on the privacy and legal issues raised by this work.

## 2      Background

### 2.1    Foundation Models

Foundation models often use a transformer architecture [3] with hundreds of billions of parameters. They are trained on large amounts of data using considerable computer resources. For example, GPT-1 was trained on 5GB, GPT-2 on 40GB, GPT-3 on 45TB and MusicGen on 20,000 hours of audio [4, 5]. After training, foundation models are given a prompt, such as a text input, and they generate an output, such as text, code or images. Some examples of foundation models are given in Table 1.

The behavior of foundation models is often fine-tuned through a combination of reinforcement learning, retrieval augmented generation (RAG) and prompt engineering. In reinforcement learning the model generates multiple different outputs, which are ranked, typically by a human, and then it is trained to generate the desired output more frequently in the future. In retrieval augmented generation a vector database is created with a set of documents that the foundation model is required to use. When the user enters a query, a vector search identifies documents that are most relevant to the user's input and these are combined with the user's query in the final prompt that is sent to the model. Prompt engineering is a variety of techniques that are used to structure prompts to get desired outputs. These include intents, roles, chains of thought and output constraints [6].

Table 1. Examples of mappings carried out by some of the current foundation models.

| Input | Output | Foundation Models |
| --- | --- | --- |
| Text | Text | GPT 3.5 Claude 3 |
| Text | Images | DALL-E, Stable Diffusion, GPT 4 |
| Text | Code | CoPilot, CodeWhisperer |
| Text | Music | MusicGen |
| Text, images | Robot actions | RT-2 |
| Text, images, DNA | Text | Med-PaLM M |
| DNA | Cellular function | Geneformer, scGPT |

## 2.2   Personality Modelling with Foundation Models

Foundation models can imitate the conversational styles of individual people. With simple prompt engineering, Chat-GPT can generate text in the style of personalities, such as Donald Trump, whose speeches and tweets were included in the original training data. More sophisticated models have been created by companies like Facebook (imitations of Snoop Dogg, Tom Brady and other celebrities) and UneeQ, who constructs digital avatars with different personalities (https://www.digitalhumans.com/). For the most part, these chatbots appear to be generated by a combination of reinforcement learning, RAG and prompt engineering. The only potential exception that we are aware of is Soul Machines (https://www.soulmachines.com/), whose website claims that their AI characters are based on a combination of LLMs and multimodal cognitive models. Details are lacking, but our best guess is that cognitive models are used to simulate the agent's state, which is combined with previous knowledge (retrieved using RAG) to build the LLM prompt. Similar work has been carried out by Park et al. [7] and Kirk et al. [8].

## 2.3   Emotions, Physiology and Decision-making

It is becoming increasingly recognized that emotions and physiological states play a central role in our decision-making and behavior. Consider a person who is sitting in front of a beef burger in a restaurant. Suppose they are experiencing hunger, and they predict that eating the burger will cause them to experience pleasurable sensations and a feeling of satiety. In this case their current and predicted physiological state explain their action of eating the burger. On the other hand, suppose that the person is feeling sick or cares deeply about animals. In this case they will not eat the burger and this behavior will, again, be explainable in terms of their current and predicted emotional and physiological states. This relationship between emotions, physiology and decision-making is nicely described by Damasio [9], whose theory of somatic markers explains how we associate positive and negative emotions with objects in our environment. Goel's [10] theory of tethered rationality is based on the idea that emotional and physiological states play a central role in the selection and initiation of behavior. A close relationship between emotions and decision making has also been demonstrated in many psychological studies [11, 12].

Within AI research, the important role that emotion plays in cognition is coming to be more widely recognized. As Pessoa puts it "Emotion is not an 'add on' that endows a robot with 'feelings' (for instance, reporting or expressing its internal state). It allows the significance of percepts, plans, and actions to be an integral part of all its computations." [13], p. 168. An AI model of a person that does not include their emotional and physiological reactions will, at best, approximate the surface level. It will not include the motivation behind their behaviors or the diversity of people's behaviors (the different burger-eating outcomes in our example).

### 2.4    Recommendation Systems

Recommendation systems typically use collaborative and content filtering. Collaborative filtering is based on similarities between users. Users rate the items (media, products, etc.) and the similarity between users is calculated on the basis of the ratings that the users give to these items. So, users might be considered similar if they give the same ratings to ten films despite there being big differences in their age and regardless of the style and contents of the films that were rated. When similar users have been identified, their ratings are used to predict the rating that a person would give to an item that they have not consumed. For example, if similar users all highly rate the film *Arrival* (2016), *Arrival* will be recommended to a person who has not seen this film before.

With content-based filtering each item is associated with a set of features. For films, these could include the genre (action, romance, documentary, etc.), actors, director, plot, etc. Each user is modelled by looking at their history for patterns in the features of items that they consume. For example, Brian might have watched a lot of *action* films starring *Dwayne Johnson*. This model of past behavior is used to recommend new content. For example, a new action film starring Dwayne Johnson would be recommended to Brian; a new romcom with Kate Winslet would not. Content-based and collaborative filtering use a variety of variety of machine learning algorithms – for example, statistics, clustering and deep neural networks - to identify similar users and recommend items based on their contents. Many recommender systems use a hybrid approach that combines content-based and collaborative filtering.

Collaborative and content-based filtering both rely on historical data from users. So they cannot provide meaningful recommendations to new users who have just joined the system. This is known as the *cold start problem*. Collaborative filtering also cannot provide good recommendations for new items that have not been rated by users.

## 3 First-person Foundation Models (FPFMs)

### 3.1 Introduction

Our hypothesis is that a foundation model that is trained from scratch on the stimuli and emotional and physiological states of a person will replicate human behavior more effectively than surface-level approximations built with LLMs, RAG, cognitive models and prompt engineering. This type of foundation model will be referred to as a *first-person foundation model* (FPFM). It could carry out the following mappings:

1. *Image/audio/text → Emotional/physiological state*. How people's emotional and physiological states change in response to different stimuli. For example, I feel sad when I look at a picture of my deceased grandmother.
2. *Emotional/physiological state → External behavior*. What people do or say when they are feeling a particular way. For example, I engage in food-seeking behavior when I am hungry.
3. *Image/audio/text + Emotional/physiological state → External behavior*. What people do or say when they are feeling a certain way and experience a particular stimuli. For example, I eat a burger when I perceive a burger in front of me and I am hungry.

### 3.2 Training Data for First-person Foundation Models

We have developed a first-person recorder that stores what a person is seeing and hearing as well as their emotional and physiological reactions to these stimuli (see Section 4). To obtain training data for a FPFM one or more people could be paid to wear this recorder for an extended period of time. This could be accomplished through companies, such as Surge AI (https://www.surgehq.ai/), who recruit people to generate training data and fine tune AI models. Another option would be to release a cheap version of the recorder and launch a marketplace where people could be paid for their recordings. People could also be motivated to use the recorder in exchange for benefits, such as life logging, personal assistance, enhanced recommendation, perfect memory, or generation of media content based on their lives.[1] This would be similar to the way in which big technology companies give us free and useful services, such as email, calendars and social networking, in exchange for access to our personal data.

### 3.3 Modelling Individuals with FPFMs

Individual FPFMs could be created from data recorded from single people. However, model training is expensive, and it would take a significant amount of time to capture enough data from one person to train a FPFM (see Section 4.2). A more practical approach would be to train a FPFM on data from multiple people and then use fine tuning and RAG to customize the model for a specific individual. Suppose we want to model

---

[1] Perfect memory and generation of media content from personal data were dramatized in the *Black Mirror* episodes The Entire History of You and Joan is Awful.

Alice, whose recordings are *not* part of the training data. A FPFM could be used to model her emotions, physiology, and behavior in the following way:

1. *Record first-person data*. Alice wears the first-person recorder for an extended period to capture her emotional and physiological responses to her environment.
2. *Fine tuning*. A local copy of a FPFM would be fine-tuned, using QLoRA [14] or a similar approach, on the user's local machine, phone or private cloud environment.[2] To begin with, historical data captured from Alice would be used to fine tune the model offline. Fine tuning would then be continued in real time as Alice wears the device. The model would predict how Alice's emotions, physiology, and behavior will change in response to visual and auditory stimuli in her environment, and these predictions would be compared with her actual emotions, physiology, and behavior. Discrepancies would be used to update the model and to monitor the performance of the model over time.
3. *RAG*. Alice's data would be stored in a vector database using a new embedding model developed for data captured by the first-person recorder. To query the FPFM, a combination of visual and audio input and emotional/physiological responses would be converted into numerical format and used to search for related data. The returned documents would be added as context to the prompt. Prompt engineering techniques could be used to constrain the model to respond like Alice, and not like one of the individuals whose data was used to train the FPFM.

### 3.4   Applications

Some potential applications of FPFMs are as follows:

- *Recommendation*. FPFMs could be used to create a new form of recommendation engine that measures the customer's actual preferences for each product and recommends the ones with the highest net valence. For example, a FPFM based on *my* emotional reactions could watch every single film and TV series on Netflix and recommend the ones that generate the most positive emotional states. This recommendation method does not depend on other users, so there would be no cold start problem when new users joined the system, or new products were added to the system.
- *Focus groups*. FPFMs based on target audiences could be used to evaluate films, products, political policies, etc. prior to their release.
- *Dialog in novels and scripts*. Current foundation models, such as GPT-4, are already being used to generate novels and scripts. FPFMs could model characters more effectively, leading to more realistic dialog. It would be possible to use FPFMs based on recordings of individual actors to generate scripts tailored to these actors. For example, a FPFM of Tom Cruise could be used to write the script for the next *Mission Impossible* movie.
- *Personal assistants*. A FPFM that understands a user's preferences could search for holidays, restaurants, etc. that appeal to the user. If a personal assistant had third-party access to the user's friends' FPFMs (see Section 3.5), then it could book group activities, such as dinner at a restaurant, that are likely to be enjoyed by all the participants (and not elicit strong negative reactions in any of them).
- *GAN systems*. A FPFM could be used as the discriminator in a generative adversarial network, providing feedback about whether text, music, images, etc. generated by a foundation model are likely to produce positive emotional responses in a specific consumer. This could be a powerful way of improving the output quality of foundation models without human feedback.
- *Dating and recruitment*. Volar (https://www.volardating.com) uses the interactions between chatbots based on two people to reduce the awkwardness of first dates. Interactions between the FPFMs of prospective partners would be a much more accurate way of evaluating the suitability of a match.[3] A similar method could be used to evaluate whether job candidates are compatible with a company's current team.

---

[2] Small FPFMs could be similar to the small language models that have been developed to run on phones, such as Phi-3.

Other applications include bereavement support, assisting dementia patients with a model of their former selves, patient models for psychologist training and phobia treatment.

### 3.5    Digital Twins and Third-party Access to FPFMs

The benefits of FPFMs could be realized by streaming first-person data to a cloud service managed by a big technology company, such as Microsoft or Amazon, who would use fine tuning and RAG to build a FPFM of the person. However, many people will be unhappy about sharing this type of data with large companies. An alternative approach would be to stream the first-person data to a local device, where it could be used to create an up-to-date digital twin of the person, which would remember everything that the person saw and heard as well as their emotional and physiological reactions. This digital twin could have access to other sources of data on the device, such as mail, photos, GPS, calendar, etc., which it could use to act as a personal assistant to the user.

To support recommendation, advertising, dating and other applications, the FPFM could be exposed to authorized third parties through a web service. To protect users' privacy, third-party access could be limited to the FPFM's emotional and physiological reactions to visual and auditory data (the first mapping in Section 3.1). Technology companies could pay a small fee to cover the costs of running the model, which would be a way in which users could benefit from money spent on advertising. Only users would have access to the behavioral output of the model (the second and third mappings in Section 3.1). The architecture of this type of application is shown in Fig. 1.

---

[3] This was dramatized in the Hang the DJ episode of *Black Mirror*.

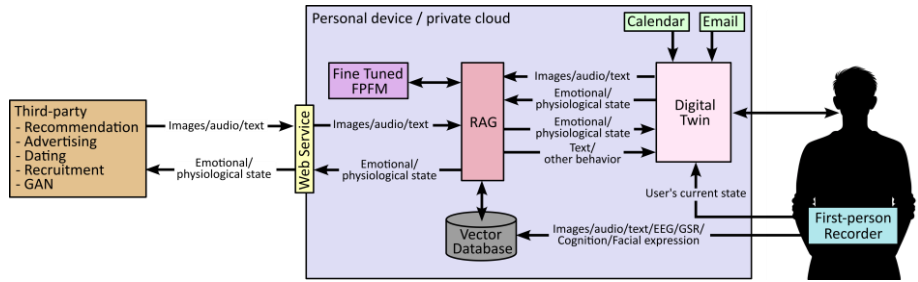

**Fig. 1.** Third party access to private FPFM. The digital twin runs in the user's private device or private cloud and has access to the user's current emotional and physiological state, as well as private data, such as calendar and email. The digital twin uses this information to act as a personal assistant – searching for holidays, clothes, jobs etc. that produce positive emotional states in the user. Third parties, such as media providers and dating applications, could be authorized to access the person's FPFM through a web service. The data returned by this service would be limited to the user's emotional and physiological reactions to images, audio, and text. This would be enough to support the most useful applications of FPFMs without exposing the user's recorded history to third parties.

## 4 First-person Recorder

We have developed a first-person recorder to capture training data for FPFMs. This records what the wearer is seeing and hearing, their emotional and physiological reactions to these stimuli and some aspects of their external behavior.

### 4.1 Hardware and Software

The recorder is based on a Raspberry Pi, worn around the user's neck, which is connected to a camera, microphone, GSR sensor and speaker. Data recorded by the Raspberry Pi is sent to a web service running on a laptop carried by the user, which has a WebSocket connection to the Epoc X EEG headset. Cloud services, such as AWS Rekognition, and the Emotiv Cortex API,[4] are used to analyze the raw data for higher level properties, such as text contents, sentiment, cognition, facial expression, and object labels. A website hosted on the laptop enables the recorder to be configured and supports playback of recorded data (see Fig. 2b). To reduce fraud a blockchain architecture is implemented that sends a hash of the data to the cloud and receives a hash of the data plus a random number known only to the cloud service, which is added to the next file in the sequence. To ensure the privacy of other people, all faces are automatically blurred during the recording process. The data is stored in JSON files; schema definitions for these files are available on the project website. Version 1.0 of the recorder is shown in Figure 2a.

---

[4] See: https://emotiv.gitbook.io/cortex-api.

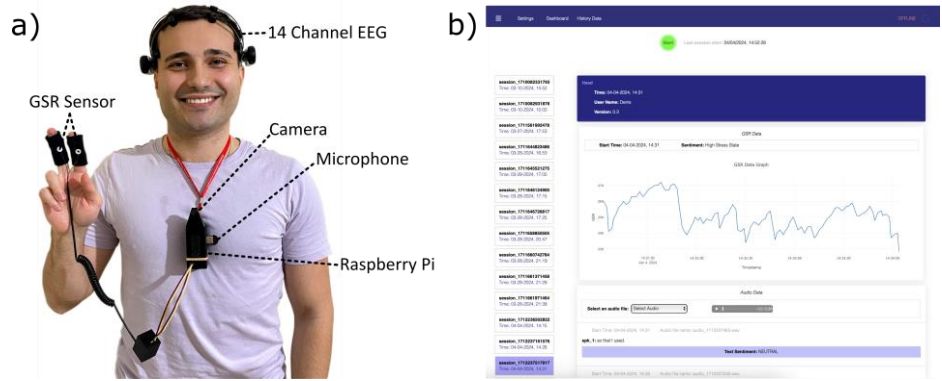

**Fig. 2.** First-person recorder. a) Hardware, including 14 channel EEG, GSR sensor, microphone and Raspberry Pi. b) Web interface that controls the recording and supports viewing of recorded data.

Code, 3D print files, JSON schemas for the data files, user manual and sample recordings are available at the project's GitHub repository: (link removed to ensure anonymity during reviewing process).

### 4.2   Recorder Data

The data captured by the recorder is summarized in Table 2. This shows that the recorder can capture ~40 GB of data (images, audio and text) in a 16-hour day. If the images, audio, raw EEG and raw GSR are excluded, this figure drops to ~1 GB per 16-hour day. Based on these figures, Table 3 gives some rough estimates of how long it would take to store enough data to train foundation models on the scale of GPT-1, GPT-2 and GPT-3. The data in Table 3 suggests that a purely text-based FPFM on the scale of GPT-2 could be created from ~50 days of data recorded from a single individual. Larger models are likely to require data recorded from several individuals.

The current version of the first-person recorder could be improved in several ways. For example, the backend could be moved to the cloud, the hardware could be upgraded, and an eye-tracking system, such as the Tobii Pro[5] could be used to capture what the user is actually looking at, instead of their general field of view. Further work is also required to integrate the recorded signals into a single picture of the emotional and physiological state of the wearer. This could be done through a machine learning approach, in which the DES data is used as the labels and a deep network is trained to automatically generate the labels from the recorded data. Since people's emotional and physiological reactions vary widely, it is likely to be necessary to calibrate the recorder for each user. EEG headsets are expensive and uncomfortable to wear, so in the future, accurate measurement of the subject's emotional and physiological states could be used to find ways of capturing approximations to this data using cheaper hardware, such as the Fitbit and Apple watch.

---

[5] See: https://www.tobii.com.

**Table 2.** Data captured by first-person recorder. Schemas for the output files and an example recording are available on the project's GitHub page. Sample rates, such as image frequency, are configurable through the web interface. The data rates are based on the test recording on the project's GitHub repository, which contains one image per second. Data rates can vary, depending on the behavior and environment of the user.

| Data | Source | Description | Rate (Kbps) |
|---|---|---|---|
| EEG | Emotiv Epoc X headset | Raw EEG data from 14 channels | 30 |
| Audio | USB microphone | Audio from microphone mounted on front of user. | 20 (mp3) |
| Images | ZeroCam | Pictures from camera mounted in front of user. | 600 (jpg) |
| GSR | Grove GSR sensor | Galvanic skin response (GSR) of user. | 0.01 (1 Hz) |
| EEG band power | Emotiv Cortex API | EEG power in the theta, alpha, beta L, beta H, and gamma bands. | 8 |
| Facial expression | Emotiv Cortex API | Eye action and expression on upper and lower face. | 4 |
| Cognition | Emotiv Cortex API | Cognitive states, including engagement, excitement, stress, relaxation, interest and focus. | 0.02 |
| Audio text | AWS Transcribe | Recorded audio is converted to text using AWS Transcribe. Speech of wearer is automatically separated from other people's speech using an audio sample generated by the user. | 0.003 |
| Speech sentiment | AWS Comprehend | Text generated by user is analyzed for sentiment (positive, negative, mixed, and neutral) using AWS Comprehend. | 0.002 |
| DES | User | Descriptive Experience Sampling (DES) is a technique in which a person describes the contents of their consciousness when prompted by a tone [15]. Key phrases, such as "Start Ziggy" and "End Ziggy", identify the start and end of a DES report. | 0.001 (per report) |
| Image text | AWS Rekognition | Text in recorded images is identified using AWS Rekognition. | 0.001 |
| Image labels | AWS Rekognition | Labels for objects in recorded images are generated using AWS Rekognition. | 2 |

**Table 3.** Time required to record enough full data or text data (without raw GSR or EEG) to train FPFMs on the scale of GPT-1, GPT-2 and GPT-3. The recording times are based on 16-hour days. The text data excludes raw GSR and raw EEG.

| Model | Training Data (GB) | Recording Time (Days, Full) | Recording time (Days, Text) |
|---|---|---|---|
| GPT-1 | 5 | 0.14 | 6.5 |
| GPT-2 | 40 | 1.1 | 52 |
| GPT-3 | 46080 | 1300 | 60000 |

## 5     Privacy and Legal Issues

The first-person recorder captures everything that the wearer is exposed to, including copyrighted books, music, and films. To reduce this problem, recordings could be automatically screened for copyrighted content or GPS could be used to switch the recorder off automatically in situations in which copyright is likely to be an issue, such as concerts and cinemas. However, the elimination of copyrighted content from the model would reduce the power of the FPFM as a recommendation engine, since it would no longer be storing the user's reactions to books, films, and music.

The automatic blurring of faces that the recorder uses to protect privacy would prevent a FPFM trained on the data from learning the strong emotional reactions that we have to familiar faces, such as friends, family, colleagues, and celebrities. To address this issue people could be asked to give their consent to have their faces recorded by the device. This could be at an individual level (I give my consent for my wife to record my face on her device) or globally (celebrities could give their consent to be recorded by anyone). Face recognition could then be used to record the faces of consenting people in an unblurred state.

Companies and government agencies with third-party access to a person's FPFM (see Section 3.5) could assess whether that person has a liking for illegal products, such as drugs, sexual attraction towards children or emotional attachment to terrorist organizations. Like any foundation model, FPFMs will frequently hallucinate. So, if a person's FPFM has strong positive responses to illegal activities, this does not prove that they have committed any actual offences or are likely to do so in the future. Great care will have to be taken with the security of FPFMs, the terms and conditions under which they are shared with third parties, and the interpretation of their outputs.

## 6     Conclusions

The central role that emotions and physiology play in the selection and initiation of behavior is widely acknowledged in neuroscience and psychology, and it is starting to be recognized in artificial intelligence research. If this interpretation of the importance of emotion is correct, foundation models that are trained on text and image data from the Internet will only be able to create surface-level approximations to the behavior of individual people.

This issue could be addressed by a new type of foundation model that maps environmental stimuli to a person's emotional and physiological states, and maps a person's emotional and physiological states to their behavior. There are many exciting applications of FPFMs, including recommendation, focus groups, dialog writing, personal assistance, GAN systems, dating and recruitment. Some of these applications could be realized by exposing a FPFM of the user's emotional and physiological reactions to trusted third parties. FPFMs could also be used to create digital twins that serve as personal assistants and are accessed privately by their owners.

To obtain training data for FPFMs we have developed a recording rig that stores what the wearer is seeing and hearing as well as their emotional and physiological reactions to their environment. Data captured by our rig could also contribute to the training of new versions of the current foundation models, such as Chat-GPT, which are experiencing a data shortfall. Data gathering and model training are expensive, so we are currently working on the launch of a start-up that could raise funds to record a substantial amount of first-person data and train a FPFM model.